\documentclass[letterpaper,10pt,conference]{ieeeconf}  
\IEEEoverridecommandlockouts
\usepackage{graphicx}
\usepackage{amsmath,amssymb}
\usepackage{booktabs}
\usepackage{array}
\usepackage{multirow}
\usepackage{cite}
\usepackage{url}
\usepackage{xcolor}
\usepackage{balance}
\usepackage{siunitx}
\usepackage{threeparttable}
\usepackage{mathtools}
\usepackage{hyperref}

\begin{document}


\title{Explainable Geospatial AI for Satellite Ground Station Siting Using LiDAR-Derived Terrain Intelligence}

\author{\authorblockN{Shohini Sarkar$^{\star}$, Smithi Mahendran, Rishi Chudasama, Varun Mannam, Arav Luthra,\\
Yuvraj Rekhi, Vivek Nadig, and Arsh Goenka}
\authorblockA{University of Maryland, College Park, MD 20742, USA}
\thanks{\raggedright
$^{\star}$Corresponding author: Shohini Sarkar,
email: \texttt{rheasarkar010@gmail.com}.
Email addresses of other authors:
Smithi Mahendran: \texttt{smithi.mahendran@gmail.com};
Rishi Chudasama: \texttt{rishiac@outlook.com};
Varun Mannam: \texttt{varunm18@terpmail.umd.edu};
Arav Luthra: \texttt{aluthra2@terpmail.umd.edu};
Yuvraj Rekhi: \texttt{yrekhi@terpmail.umd.edu};
Vivek Nadig: \texttt{nadig.vivek@gmail.com};
Arsh Goenka: \texttt{arshgoenkarulz@gmail.com}.}
}

\maketitle

\begin{abstract}
Representative clutter height (RCH) is a key parameter in radio propagation and interference analysis because it captures the dominant height of local obstructions that drive terminal clutter loss. Current practice often relies on fixed clutter heights assigned to land use classes in Recommendation ITU-R P.452-18, but this misses within class variation and can lead to conservative exclusion zones and poor site ranking for low Earth orbit ground station siting and spectrum coordination. We present an interpretable, globally deployable machine learning framework for predicting RCH from open geospatial data. The model is trained using LiDAR derived labels from the U.S. Geological Survey 3D Elevation Program and inference time features from global land-cover, terrain, demographic, thermal, and optical remote sensing products. We define RCH using a robust 75th percentile clutter height statistic, evaluate multiple regressors, and select LightGBM for its accuracy, efficiency, and compatibility with feature attribution analysis. The final model achieves a mean absolute error of \SI{1.79}{m} and an $R^2=0.765$, reducing absolute error by more than 60\% relative to the ITU baseline. Beyond aggregate fit, we evaluate domain facing criteria relevant to RF planning, including meter scale error, tolerance band accuracy, over and under estimation tails, agreement with ITU clutter height regimes, and SHAP-based physical plausibility. SHAP identifies tree canopy cover, land-cover semantics, and spectral reflectance as the most influential predictors. Studies on segmentation derived features, non-forest ablations, and land-cover matched international validation show that open geospatial data can improve clutter modeling at scale without sacrificing interpretability or deployability.
\end{abstract}


\noindent\textbf{Demo Video}: \url{https://youtu.be/H3WRpq-XwWA}

\section{Introduction}
The ongoing expansion of satellite broadband, Earth-observation, and Internet-of-Things infrastructures has made the selection and coordination of ground-station sites a high-stakes technical problem. As non-geostationary constellations scale, operators must identify geographically distributed sites that are not only topographically viable, but also compatible with surrounding clutter environments that shape clutter loss, link margins, and interference risk. In parallel, terrestrial networks continue to densify, creating a planning context in which inaccurate environmental assumptions can translate into either under-protected links or overly conservative coordination boundaries.

\begin{figure*}[t]
\centering
    \includegraphics[width=2\columnwidth]{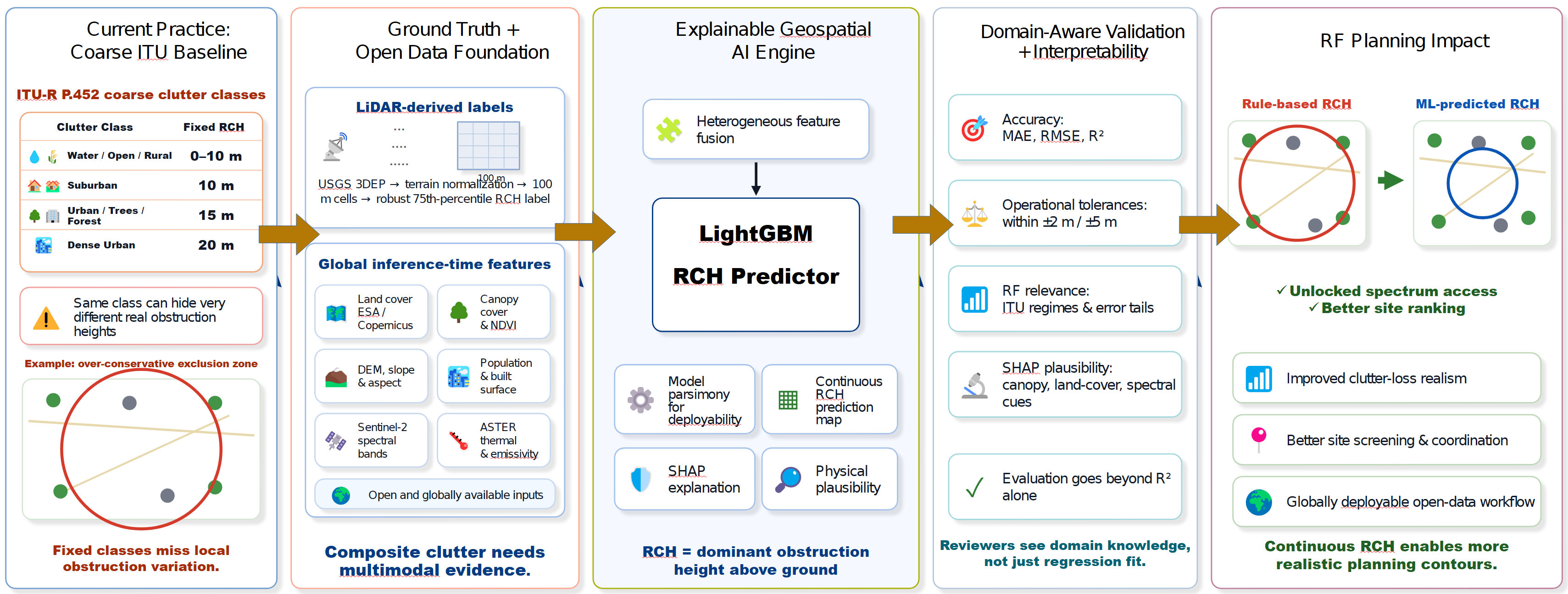}
\caption{Explainable geospatial AI converts open remote-sensing data and LiDAR supervision into continuous, physically interpretable clutter estimates, replacing coarse categorical defaults and enabling more accurate satellite ground-station siting and spectrum coordination.}
\label{fig:sys}
\end{figure*}

A central parameter in these studies is \emph{Representative Clutter Height} (RCH), which summarizes the effective height of local obstructions above the underlying terrain. In current engineering practice, Recommendation ITU-R P.452-18 maps broad categorical labels such as suburban, urban, forest, and dense urban to fixed clutter heights \cite{itu} (Table~\ref{tab:itu_clutter}). Recommendation ITU-R P.2108-1 likewise formalizes terminal clutter loss as a core ingredient in propagation analysis \cite{p2108}. These standards remain operationally useful, but their default values are intentionally coarse. They do not account for the fact that two sites with the same nominal land-use class may have very different mixtures of canopy density, building morphology, built-up intensity, and terrain structure.

This gap matters from both a technology and business perspective. For technology, fixed-category clutter assumptions limit the fidelity of interference studies, weaken site prioritization, and obscure transitional environments such as suburban--rural boundaries and mixed built-vegetated areas. From a business standpoint, conservative clutter assumptions can expand exclusion zones, reduce the number of feasible sites, and increase the survey burden required to validate candidate deployments. A globally deployable, data-driven RCH model therefore offers value not merely as a prediction exercise, but as a planning tool that can improve screening accuracy and reduce uncertainty early in the siting workflow.

Recent remote-sensing research has shown that structural height can be inferred from globally available geospatial products, including optical imagery, building footprints, and spaceborne LiDAR \cite{li,stipek,potapov,frantz,shi}. However, most prior work focuses on isolated structure classes such as building height or canopy height rather than the composite obstruction height relevant for path-specific radio engineering. Our goal is different: we seek a directly useful, globally generalizable, and interpretable estimate of RCH itself.

The paper makes five main contributions (Figure~\ref{fig:sys}, Figure~\ref{fig:pipeline}):
\begin{itemize}
    \item We define a practical supervised learning formulation for RCH prediction that is aligned with propagation engineering and uses LiDAR-derived labels from USGS 3DEP \cite{usgs3dep,pdal}.
    \item We construct a globally deployable feature stack composed entirely of open geospatial products, spanning land cover, terrain, population, thermal variables, canopy cover, and multispectral reflectance.
    \item We show that LightGBM \cite{lightgbm} substantially outperforms fixed ITU-R clutter defaults, achieving \SI{1.79}{m} MAE and $R^2=0.765$ on held-out U.S. data.
    \item We provide mathematical framing for robust target generation, gradient-boosted regression, and SHAP-based explanation \cite{shap}, enabling a more transparent interpretation of the learned mapping.
    \item We analyze deployment-oriented questions including global transferability, segmentation-feature utility, robustness outside forested environments, and the operational implications of improved RCH estimation for siting and spectrum coordination.
    \item We add a domain-informed evaluation of RCH predictions beyond $R^2$, including absolute height error in meters, tolerance-band accuracy, large-error tail rates, ITU clutter-regime agreement, and feature-plausibility
checks grounded in RF propagation and ground-station-siting requirements.
\end{itemize}

\begin{table}[!t]
\caption{Default Representative Clutter Height Values from ITU-R P.452-18}
\label{tab:itu_clutter}
\centering
\small
\setlength{\tabcolsep}{5pt}
\begin{tabular}{>{\raggedright\arraybackslash}p{0.58\columnwidth}c}
\toprule
Clutter category & RCH (m) \\
\midrule
Water / Sea / Open / Rural & 0--10 \\
Suburban & 10 \\
Urban / Trees / Forest & 15 \\
Dense urban & 20 \\
\bottomrule
\end{tabular}
\end{table}

\section{Related Work and Problem Framing}
Recommendation ITU-R P.452-18 remains a standard engineering reference for interference evaluation between terrestrial stations and therefore underpins many satellite and shared-spectrum studies \cite{itu}. Recommendation ITU-R P.2108-1 similarly emphasizes the practical importance of clutter loss in propagation workflows \cite{p2108}. These standards are not the problem; rather, the problem is that the available clutter inputs are often static and low resolution compared to the environmental heterogeneity of real deployment sites.

In parallel, machine learning for remote sensing has matured rapidly. Li \emph{et al.} \cite{li} and Stipek and Goodchild \cite{stipek} estimate large-scale building morphology using interpretable machine-learning pipelines. Potapov \emph{et al.} \cite{potapov} combine GEDI and Landsat observations for canopy-height mapping, while Frantz \emph{et al.} \cite{frantz} and Shi \emph{et al.} \cite{shi} use Sentinel imagery and auxiliary map layers to infer building height across large geographies. These works demonstrate that height can be inferred from open or widely accessible Earth-observation features, but they do not directly predict the composite clutter quantity used in RF planning.

The distinction is important. For radio engineering, the relevant target is neither building height nor canopy height in isolation; it is the dominant obstruction environment experienced by a path endpoint. In a suburban site, for example, mature tree crowns, low-rise buildings, local topography, and mixed built-surface intensity may all shape the effective clutter. This motivates a composite-label approach rather than separate single-phenomenon models.

We therefore frame RCH prediction as a supervised regression problem over a heterogeneous geospatial feature vector $\mathbf{x}_i \in \mathbb{R}^d$ associated with each spatial cell $i$, with target $y_i \in \mathbb{R}_{\ge 0}$ denoting clutter height above local ground. The desired predictor $f(\mathbf{x}_i)$ should satisfy four criteria: (i) high predictive accuracy relative to fixed engineering defaults, (ii) global deployability using open data, (iii) robustness across mixed vegetative and anthropogenic environments, and (iv) interpretability sufficient for engineering review and deployment confidence.

\begin{figure}[t]
\centering
    \includegraphics[width=1\columnwidth]{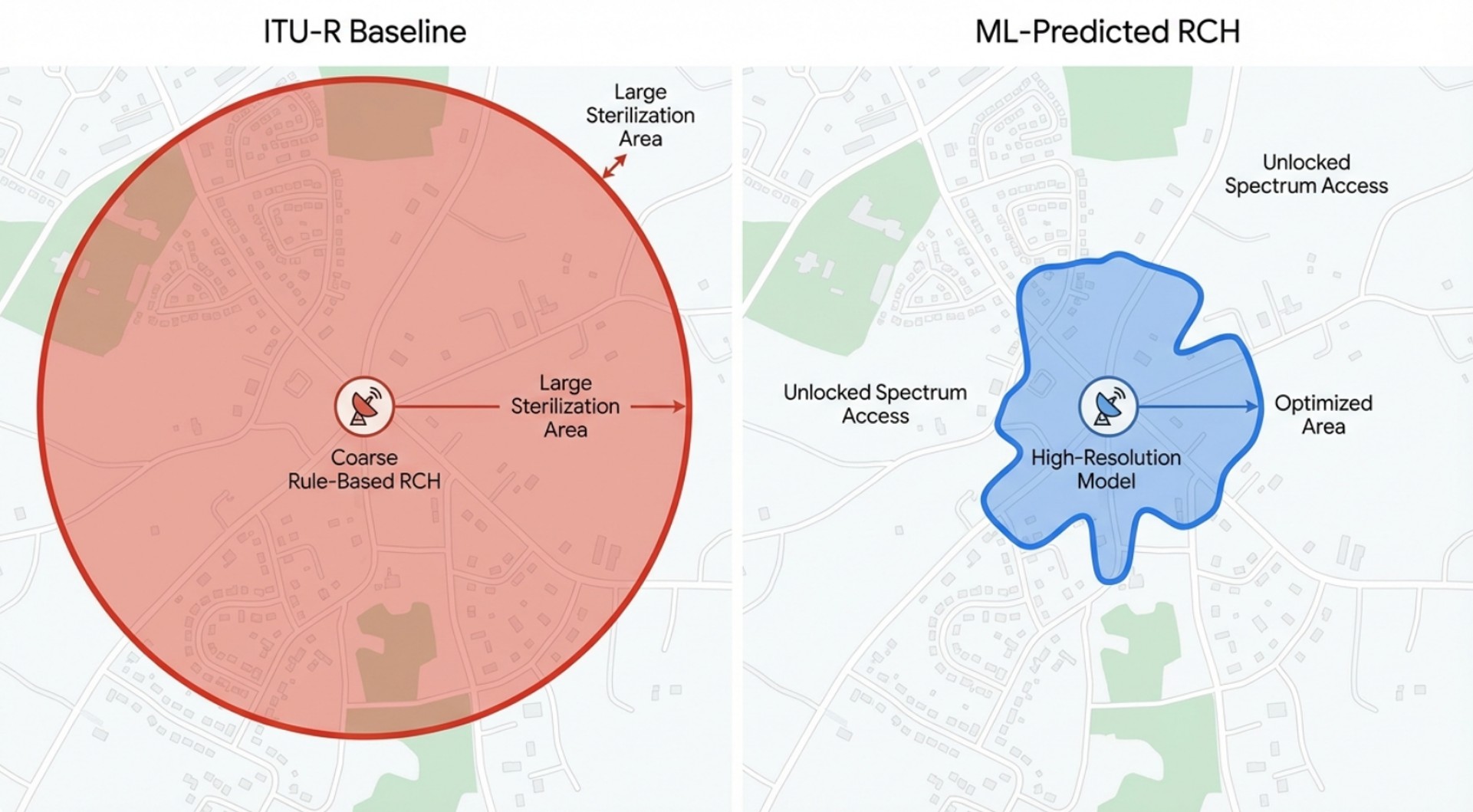}
\caption{Conceptual comparison between coarse rule-based RCH defaults and high-resolution ML-predicted RCH for spectrum coordination. The figure illustrates how more accurate clutter modeling can shrink sterilization zones and unlock additional shared-spectrum access.}
\label{fig:problem}
\end{figure}

\section{Data Foundation and Target Construction}
\subsection{Ground-Truth Derivation from LiDAR}
\begin{figure}[!t]
\centering
\includegraphics[width=\columnwidth]{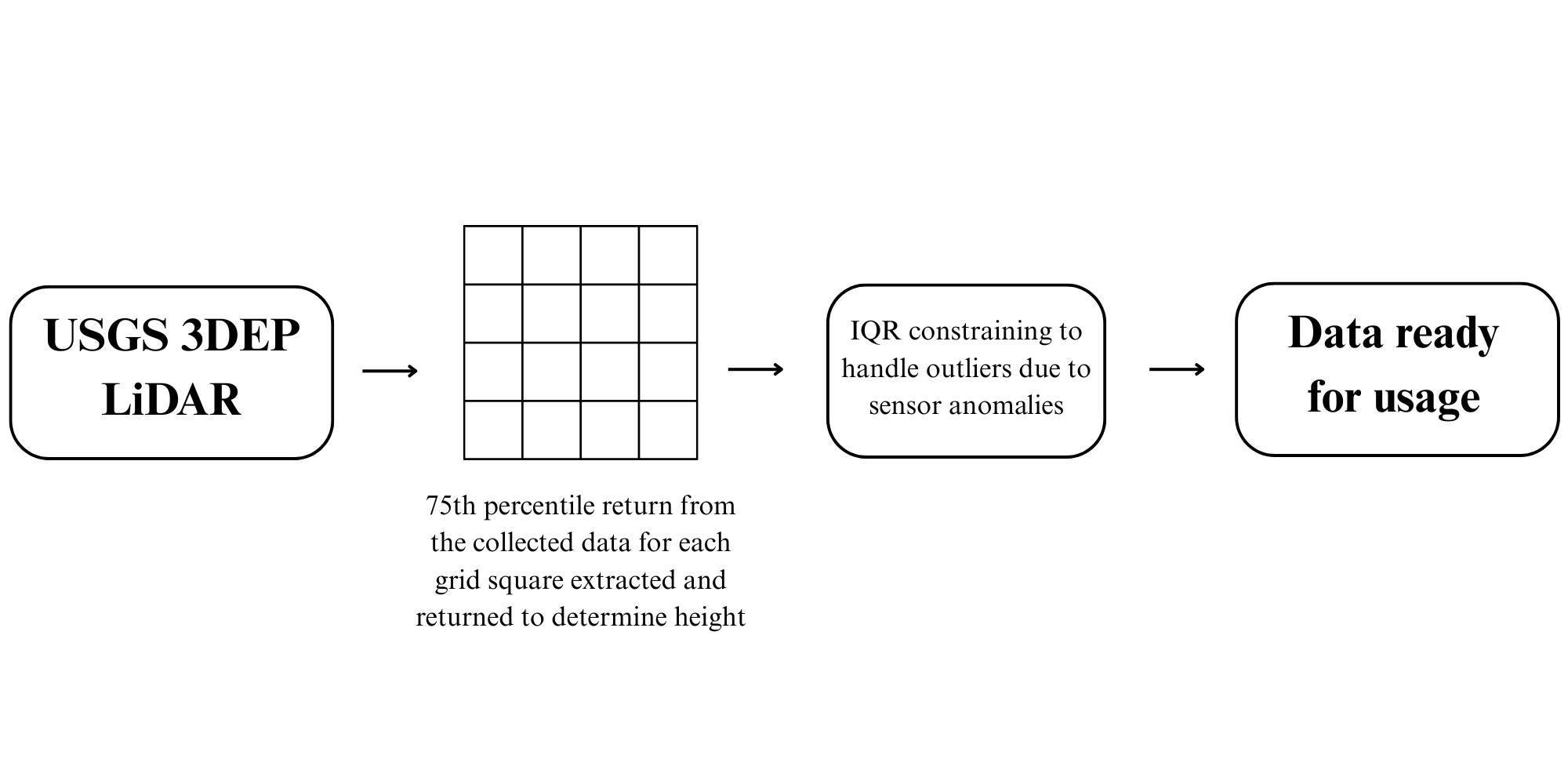}
\caption{Label-generation and modeling workflow. LiDAR point clouds from USGS 3DEP are terrain-normalized and aggregated to a \SI{100}{m} grid, a robust 75th-percentile clutter-height label is derived per cell, outlier filtering is applied, and open geospatial features are then joined for model training and inference.}
\label{fig:pipeline}
\end{figure}

A major challenge in clutter modeling is the scarcity of large, geographically diverse, and reliable three-dimensional ground truth. We address this by using the U.S. Geological Survey 3D Elevation Program (3DEP) (Figure~\ref{fig:pipeline}), which provides extensive aerial LiDAR coverage across the contiguous United States \cite{usgs3dep}. LiDAR point clouds are processed using PDAL \cite{pdal} and aggregated to a \SI{100}{m} grid to align with the spatial granularity of globally available geospatial predictors (Figure~\ref{fig:pipeline}).

Within each grid cell $i$, let $\mathcal{H}_i = \{h_{i1}, h_{i2}, \dots, h_{in_i}\}$ denote the set of above-ground return heights after terrain normalization. Rather than using the maximum return, which is highly sensitive to sparse artifacts and isolated structures, or the mean, which can understate dominant clutter in heterogeneous scenes, we define the label as the empirical 75th percentile:
\begin{equation}
    y_i = Q_{0.75}(\mathcal{H}_i),
    \label{eq:target_quantile}
\end{equation}
where $Q_{0.75}(\cdot)$ is the sample quantile operator. This choice provides a physically motivated compromise: it captures dominant obstructions that meaningfully affect clutter loss while avoiding undue sensitivity to outlier returns. Cells with unstable distributions are filtered using an interquartile-range criterion, and canopy-height cues are used as a complementary consistency signal in heavily vegetated regions.

\subsection{Feature Stack and Physical Rationale}
The final model uses a deliberately heterogeneous feature set because RCH is a composite environmental quantity (Table~\ref{tab:feature_summary}). A single modality cannot adequately capture the diversity of clutter drivers across dense urban cores, forested terrain, peri-urban corridors, and sparsely populated regions. Table~\ref{tab:feature_summary} summarizes the main feature groups.

\begin{table}[!t]
\caption{Summary of Inference-Time Input Features}
\label{tab:feature_summary}
\centering
\footnotesize
\setlength{\tabcolsep}{4pt}
\begin{tabular}{>{\raggedright\arraybackslash}p{0.35\columnwidth} >{\raggedright\arraybackslash}p{0.56\columnwidth}}
\toprule
Feature group & Inputs \\
\midrule
Terrain and elevation & Copernicus DEM, SRTM DEM, slope, aspect \\
Vegetation structure & Tree canopy cover, NDVI \\
Anthropogenic intensity & WorldPop population density, GHSL built surface \\
Land-cover semantics & ESA WorldCover, Copernicus land cover, JRC forest type \\
Spectral reflectance & Sentinel-2 B2, B3, B4, B8 \\
Thermal / emissivity context & ASTER emissivity bands 10--14 and temperature \\
Auxiliary topography & ASTER DEM \\
\bottomrule
\end{tabular}
\end{table}

The physics behind these choices is intuitive. Tree canopy and vegetation indices help explain natural clutter, built-surface and population variables correlate with the density of human-made obstructions, land-cover products encode semantic context, and terrain features capture the local morphology that shapes both obstruction patterns and radio paths. Spectral and thermal bands act as indirect descriptors of surface materials, moisture, and vegetation state, allowing the model to exploit nonlinear cues not captured by categorical maps alone.

\begin{figure}[t]
\centering
    \includegraphics[width=1\columnwidth]{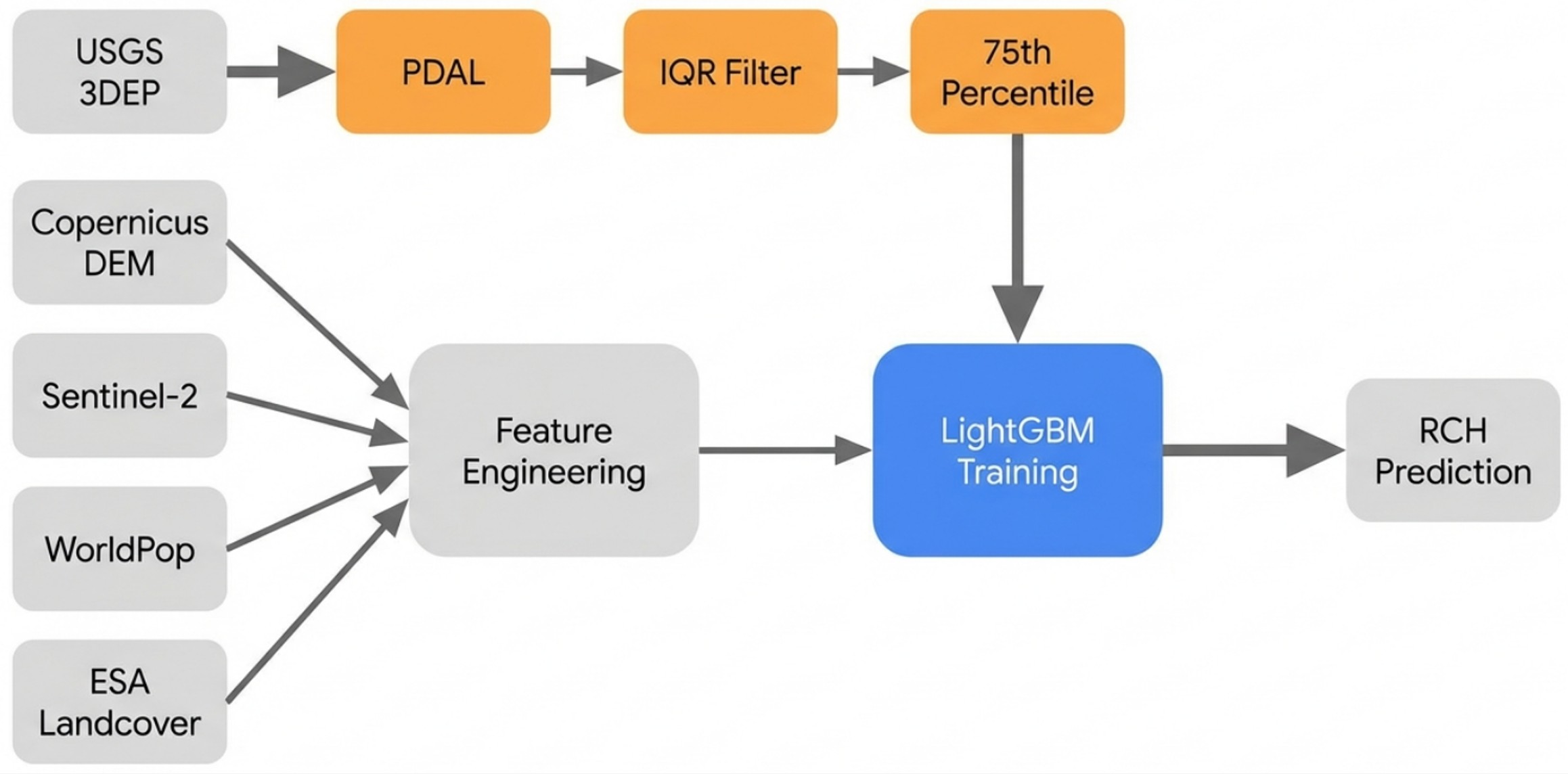}
\caption{Data-processing pipeline for ML-based clutter prediction. LiDAR-derived training labels are fused with globally available geospatial inputs for model training and deployment.}
\label{fig:dataproc}
\end{figure}

\begin{figure*}[]
\centering
\includegraphics[width=2\columnwidth]{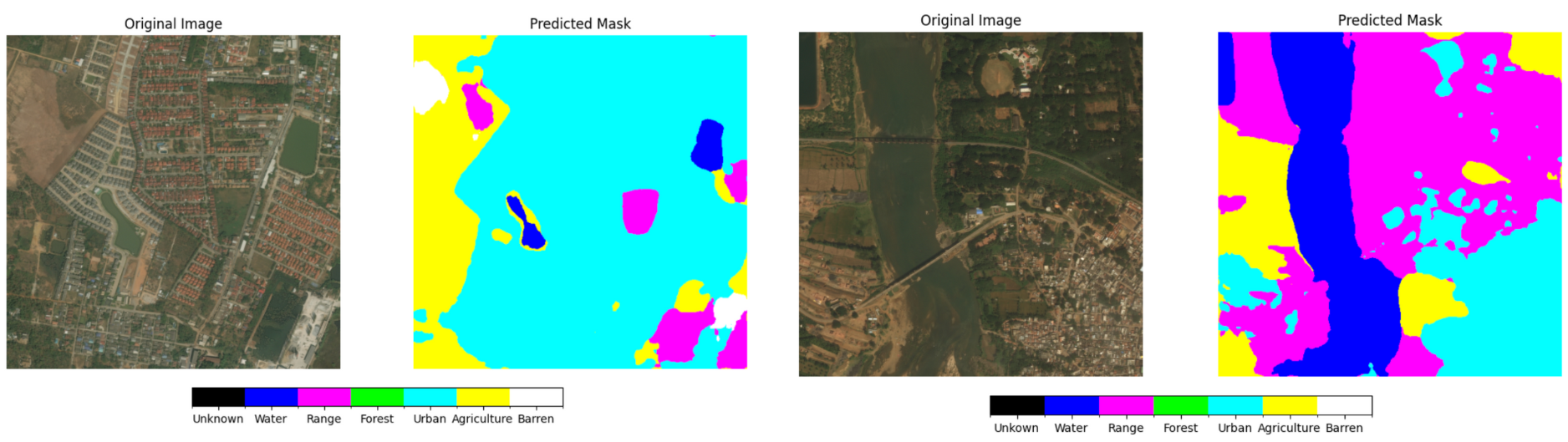}
\caption{Representative semantic-segmentation outputs explored as candidate auxiliary features. Although visually informative, segmentation-derived area fractions contributed little additional predictive power beyond existing global land-cover products at the \SI{100}{m} modeling scale.}
\label{fig:segmentation}
\end{figure*}

\section{Mathematical and Algorithmic Framework}
\subsection{Propagation Context and Learning Objective}
At a high level, path loss near a cluttered terminal can be decomposed schematically as
\begin{equation}
    L_{\text{tot}} = L_{\text{fs}} + L_{\text{atm}} + L_{\text{diff}} + L_{\text{clutter}},
    \label{eq:pathloss_decomp}
\end{equation}
where $L_{\text{fs}}$ is free-space loss, $L_{\text{atm}}$ captures atmospheric effects, $L_{\text{diff}}$ represents diffraction over terrain, and $L_{\text{clutter}}$ is the additional attenuation due to terminal-side obstructions. The exact propagation computation depends on the standard and scenario \cite{itu,p2108}, but in all cases the clutter term is highly sensitive to the assumed local obstruction height. Improving RCH therefore improves the realism of the clutter component provided to a broader propagation workflow.

We model RCH prediction as supervised regression. Given training pairs $\{(\mathbf{x}_i, y_i)\}_{i=1}^N$, the goal is to learn a predictor $f$ minimizing
\begin{equation}
    \min_f \; \sum_{i=1}^N \ell\big(y_i, f(\mathbf{x}_i)\big) + \lambda \, \Omega(f),
    \label{eq:generic_obj}
\end{equation}
where $\ell$ is a pointwise regression loss and $\Omega$ is a complexity penalty controlling overfitting. The complexity penalty is included not only as a statistical regularizer, but also to reflect domain constraints identified during engineering review. For satellite ground-station screening and spectrum coordination, domain experts require a model that is auditable, reproducible with globally available inputs, and stable under modest changes in geography or feature availability. Excessive model complexity can improve validation fit while increasing the risk of brittle site rankings, hidden dependence on local
spatial proxies, and reduced confidence during RF engineering review. We therefore treat model parsimony as an operational requirement: additional model structure or auxiliary features are retained only when they improve absolute height error or domain-facing validation checks without reducing interpretability or global deployability.

\subsection{Why LightGBM}
We evaluated multilayer perceptrons, Random Forest, XGBoost, and LightGBM. For this problem, the feature space is heterogeneous, partly tabular, and rich in nonlinear interactions but not so high-dimensional that deep representation learning is necessary. Gradient-boosted trees are therefore a natural fit. LightGBM was selected because it provided the strongest balance between accuracy, computational efficiency, and interpretability.

In boosting iteration $t$, the model adds a regression tree $f_t$ to the current predictor:
\begin{equation}
    \hat{y}_i^{(t)} = \hat{y}_i^{(t-1)} + f_t(\mathbf{x}_i).
\end{equation}
Following the standard boosting framework, the stagewise objective can be written as
\begin{equation}
    \mathcal{L}^{(t)} = \sum_{i=1}^N \ell\big(y_i, \hat{y}_i^{(t-1)} + f_t(\mathbf{x}_i)\big) + \Omega(f_t),
    \label{eq:lgbm_obj}
\end{equation}
with the regularization term on a tree typically expressed as
\begin{equation}
    \Omega(f_t) = \gamma T + \frac{1}{2}\lambda \sum_{j=1}^{T} w_j^2,
\end{equation}
where $T$ is the number of leaves and $w_j$ are leaf weights. In practice, LightGBM's histogram-based training, leaf-wise tree growth, and efficient handling of mixed feature types make it well suited to geospatial regression on large tabular datasets \cite{lightgbm}.

\subsection{Feature Attribution with SHAP}
Predictive performance alone is insufficient for deployment in an engineering setting. We therefore use SHAP to interpret the learned model \cite{shap}. For an instance $\mathbf{x}$, the prediction can be decomposed as
\begin{equation}
    f(\mathbf{x}) = \phi_0 + \sum_{k=1}^{d} \phi_k,
    \label{eq:shap_additive}
\end{equation}
where $\phi_0$ is the expected model output over the background distribution and $\phi_k$ is the contribution of feature $k$. Formally, the Shapley value for feature $k$ is
\begin{equation}
\phi_k = \sum_{S \subseteq F \setminus \{k\}} \frac{|S|!(|F|-|S|-1)!}{|F|!} \Big[v(S \cup \{k\}) - v(S)\Big],
\label{eq:shap_formula}
\end{equation}
where $F$ is the full feature set and $v(S)$ is the expected model value conditioned on a coalition $S$. This formulation is attractive because it satisfies local accuracy, consistency, and missingness, allowing us to determine whether the model is using physically sensible cues or merely exploiting geographic shortcuts.

\section{Experimental Design}
Training uses more than 50{,}000 labeled \SI{100}{m} cells sampled across the contiguous United States. We employ five-fold cross-validation with geographically shuffled splits to reduce memorization of local spatial patterns while still exposing the model to broad environmental diversity. 

Hyperparameters are selected via randomized search over tree depth, number of leaves, learning rate, estimators, and minimum leaf samples. Although $R^2$ is reported for comparability with standard regression studies, model selection is not based on $R^2$ alone. We also evaluate candidate models using deployment-facing criteria: mean absolute error in meters, the fraction of predictions within operational tolerance bands, asymmetric large-error rates, error behavior by ITU clutter regime, SHAP-based physical plausibility,
and the complexity of the required inference stack. Candidate configurations that increased statistical fit only marginally but added non-deployable features, unstable attribution patterns, or larger error tails were not preferred.

The final LightGBM configuration uses a learning rate of 0.1, 140 estimators, maximum depth of 8, 20 leaves per tree, and a minimum of 20 samples per leaf. These settings provided the best trade-off between flexibility and generalization on held-out samples. Performance is reported using MAE, root mean squared error (RMSE), and coefficient of determination $R^2$.

We compare against the ITU-R P.452-18 baseline by mapping each test cell to its corresponding categorical clutter class and assigning the associated default height from Table~\ref{tab:itu_clutter}. This baseline is intentionally simple, but it mirrors the type of coarse assumption still used in engineering workflows when high-resolution clutter data are unavailable.

\section{Results}
\subsection{Overall Predictive Performance}
\begin{figure}[!t]
\centering
\includegraphics[width=\columnwidth]{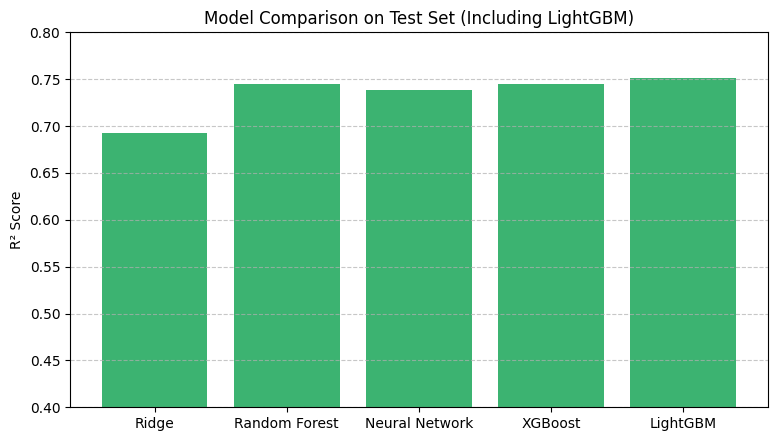}
\caption{Comparison of candidate regressors. LightGBM delivers the strongest overall balance, achieving the best $R^2$ and lowest MAE among the evaluated models.}
\label{fig:results}
\end{figure}

The proposed model substantially outperforms the engineering baseline (Figure~\ref{fig:results}). Table~\ref{tab:main_results} summarizes the primary results. LightGBM achieves an MAE of \SI{1.79}{m} and $R^2=0.765$, compared with MAE of \SI{4.67}{m} and $R^2=0.412$ for the ITU baseline. This corresponds to more than 60\% lower absolute error and a large increase in explained variance. In practical terms, the learned model captures local clutter variation that is invisible to broad categorical defaults.

\begin{table}[!t]
\caption{Main Predictive Results on Held-Out U.S. Test Data}
\label{tab:main_results}
\centering
\small
\setlength{\tabcolsep}{5pt}
\begin{tabular}{lccc}
\toprule
Method & MAE (m) & RMSE (m) & $R^2$ \\
\midrule
ITU-R P.452-18 default heights & 4.67 & --- & 0.412 \\
LightGBM (final model) & 1.79 & --- & 0.765 \\
LightGBM without forest features & 0.87 & 2.09 & 0.617 \\
\bottomrule
\end{tabular}
\end{table}

\subsection{Interpretability and Learned Physical Structure}
\begin{figure}[!t]
\centering
\includegraphics[width=\columnwidth]{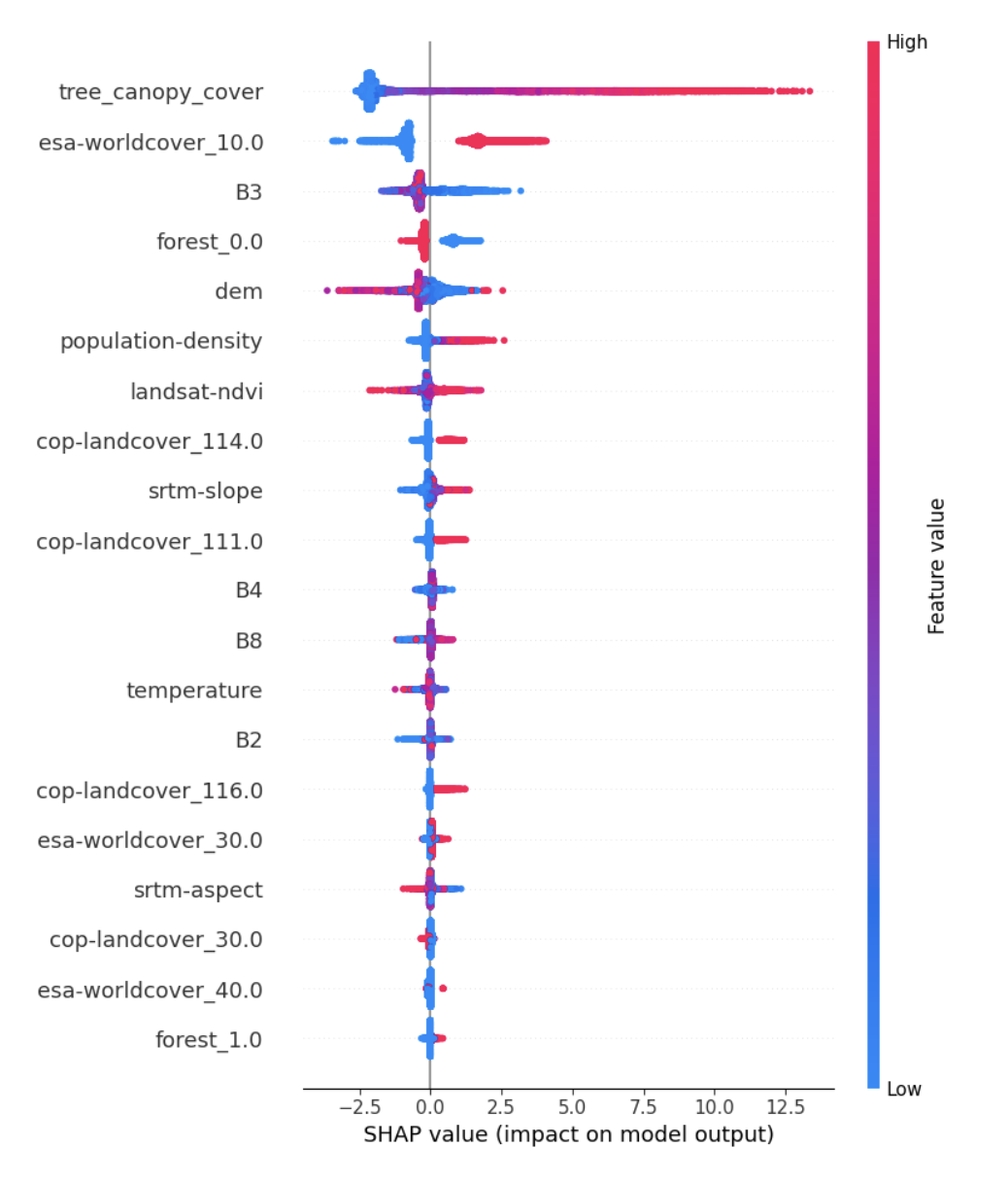}
\caption{SHAP summary for the final LightGBM model. The ordering and directionality of the top features are physically plausible: high canopy cover, informative land-cover classes, and selected spectral bands strongly influence elevated RCH predictions.}
\label{fig:shap}
\end{figure}

The SHAP summary in Fig.~\ref{fig:shap} provides an important sanity check on the model. Tree canopy cover is the strongest predictor, which is physically sensible because tall, dense vegetation directly contributes to clutter height. Land-cover classes rank next, showing that semantic context remains useful even after continuous features are introduced. Sentinel-2 spectral bands are also highly informative, indicating that reflectance patterns capture structural cues correlated with both vegetation density and built morphology. Population density, built surface, and terrain variables provide complementary information in urban and mixed settings.

\subsection{Segmentation as an Auxiliary Signal}
We also investigated whether a higher-resolution semantic-segmentation pipeline would improve performance. A DeepLabV3+ model with a transformer-based encoder was trained to derive fractional class-area features from image patches. Figure~\ref{fig:segmentation} illustrates representative segmentation outputs. While these masks were visually reasonable, their derived features had negligible SHAP importance and did not materially improve regression performance.

This result is itself useful. It suggests that at the \SI{100}{m} scale, contemporary global products such as ESA WorldCover and Copernicus already provide enough semantic information for clutter modeling, reducing the need for a more complex deployment stack. In other words, the winning design is not the most complicated one; it is the one that best balances predictive value, global deployability, and engineering simplicity.

This finding also supports the use of a model-complexity penalty from a domain perspective. Expert review of the siting workflow indicated that an additional segmentation model would introduce extra training data needs, maintenance burden, inference-time dependencies, and possible failure modes under international transfer. Because the segmentation-derived features did not materially improve the domain-facing validation metrics or the physical plausibility of the SHAP rankings, we exclude them from the final pipeline. Thus, the selected model is intentionally parsimonious: it retains the features and complexity needed for accurate RCH estimation while avoiding components that are difficult to audit or deploy globally.

\subsection{Global Generalization}
\begin{figure}[!t]
\centering
\includegraphics[width=\columnwidth]{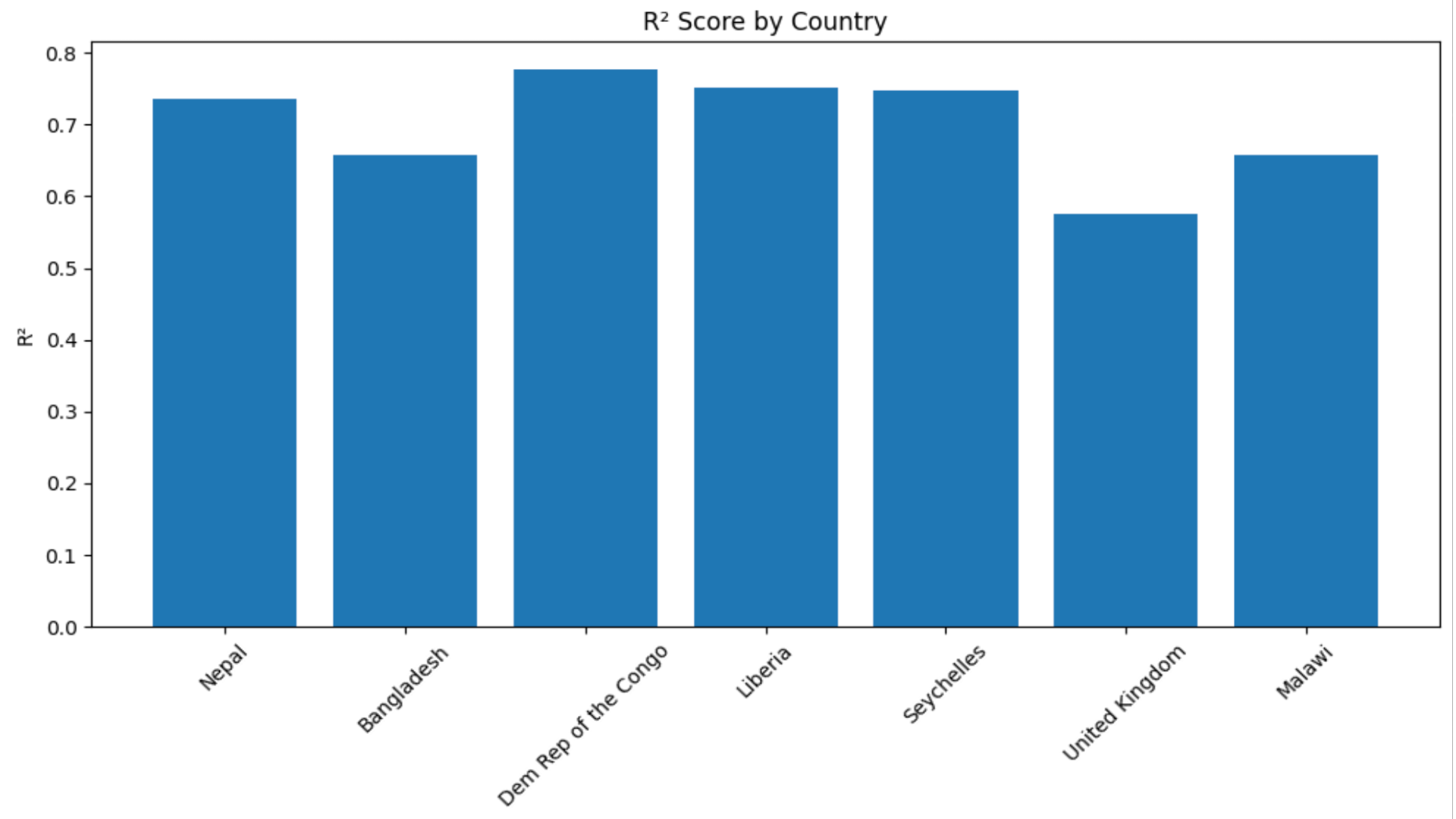}
\caption{Landcover-matched generalization performance for several countries. The model consistently exceeds the ITU baseline, suggesting useful transferability despite U.S.-only supervised training.}
\label{fig:global_generalization}
\end{figure}

A critical question is whether a model trained on U.S. LiDAR-derived labels can transfer outside the United States. Since direct international LiDAR validation was not available at comparable scale, we use a landcover-matched protocol: for each target country, Google Earth Engine is used to sample the local land-cover distribution, and a matched subset is drawn from the U.S. training distribution. Figure~\ref{fig:global_generalization} shows that $R^2$ ranges from roughly 0.58 to 0.78 across the tested countries, consistently outperforming the ITU baseline.

This is not a substitute for true overseas labeled evaluation, but it is strong preliminary evidence that the model is learning transferable environmental relationships rather than narrowly memorizing U.S.-specific geography. This finding is central to the deployment case for the method, because inference-time features are globally available even though direct LiDAR labels are not.

\begin{figure*}[t]
\centering
\includegraphics[width=2\columnwidth]{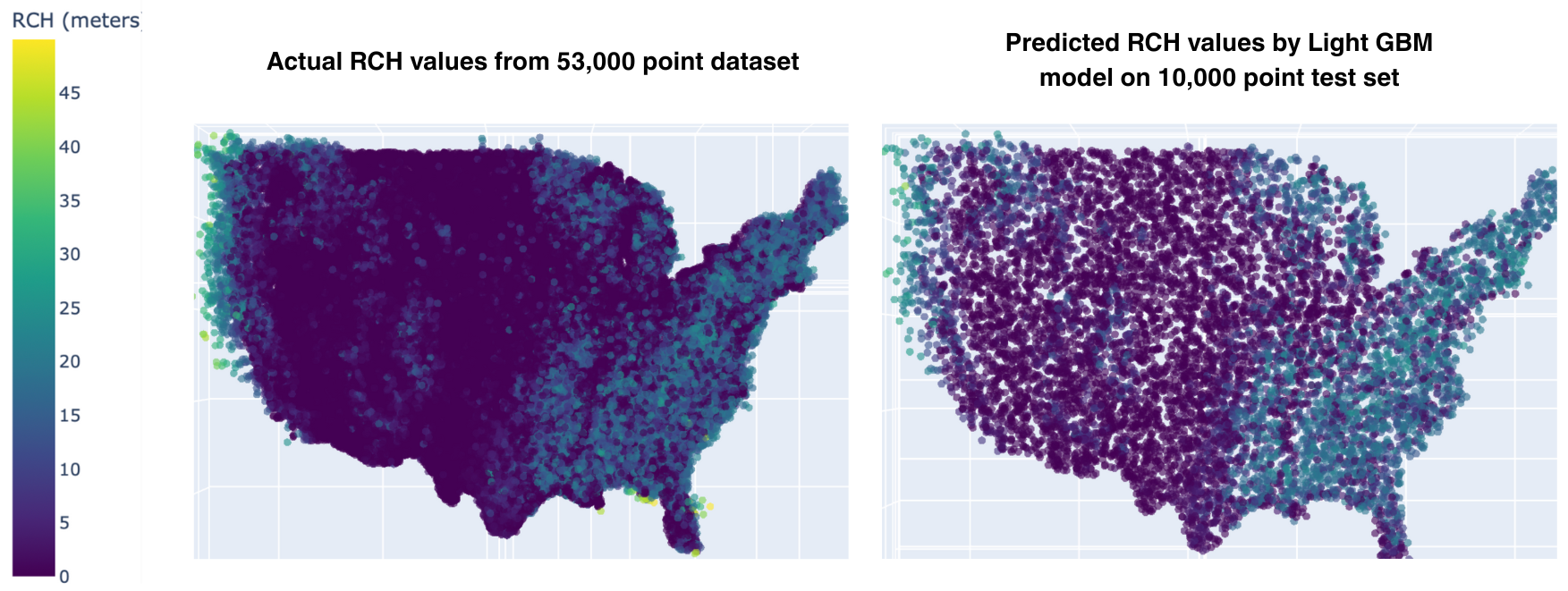}
\caption{Geographic distribution of training samples and LightGBM predictions on the held-out set. High predicted RCH values cluster in forested and structurally dense regions, while lower values are more common in open terrain.}
\label{fig:geographic_distribution}
\end{figure*}

\subsection{Geographic Patterns and Robustness}
Figure~\ref{fig:geographic_distribution} shows the geographic distribution of training data and held-out predictions. Higher RCH estimates align qualitatively with heavily forested or structurally dense regions such as the Pacific Northwest and Appalachia, while lower values occur more often in open plains and arid southwestern areas. These patterns are consistent with known environmental structure and reinforce the interpretation that the model is capturing meaningful clutter regimes.

Because canopy cover is the strongest SHAP feature, we also test a non-forest ablation in which forest- and canopy-related variables are removed. Even under this constraint, the model retains substantial predictive value with $R^2=0.617$. This indicates that NDVI, land-cover products, topographic context, built-surface measures, and demographic signals together remain informative for urban and sparsely vegetated clutter estimation.

\section{Discussion: Technology, Business, and Deployment Implications}
The results suggest that RCH can be treated as a modern geospatial inference problem rather than a fixed lookup-table parameter. This shift has several implications.

\subsection{Technology Impact}
First, the proposed approach replaces static land-use assignments with a continuous estimate grounded in multimodal evidence. That matters because clutter is inherently composite: it emerges from the interaction of trees, built structures, land-cover semantics, and topographic context. A model that uses these cues jointly is better aligned with physical reality than any single-source proxy.

Second, the learned model is interpretable enough to support engineering review. In operational settings, decision-makers often need to know \emph{why} a site received a high clutter estimate. SHAP provides that bridge, helping users verify that high values are being driven by expected environmental signals rather than opaque artifacts. This is especially valuable in satellite coordination workflows where model outputs may influence exclusion boundaries, site ranking, or requests for further field validation.

Third, the framework is globally deployable because all inference-time features are drawn from open data products. That design choice is intentional. A model that requires proprietary footprints or region-specific ancillary layers may benchmark well but fail to support real global screening. Here, the emphasis is on a practical engineering tool that can be run anywhere with a consistent input stack.

\subsection{Business Motivation and Operational Value}
From a business perspective, improved clutter modeling can create value at multiple stages of network planning. During early screening, it can reduce false positives and false negatives in candidate-site evaluation. During coordination studies, it can replace blunt categorical assumptions with site-specific estimates, reducing the need to treat heterogeneous regions as if they were uniformly obstructed. For large satellite operators or ground-segment providers evaluating many candidate sites, even modest improvements in filtering accuracy can reduce survey effort, accelerate deployment timelines, and improve capital allocation.

There is also a strategic value in better understanding \emph{where} the baseline fails. The strongest gains appear in transitional and mixed environments where a fixed categorical rule is least representative. Those are often the very regions where business decisions are difficult: fast-growing suburban corridors, semi-forested logistics zones, and mixed-use peripheries that are attractive for deployment but hard to characterize quickly. A continuous, interpretable clutter model is therefore not only more accurate on average; it is more useful where planning uncertainty is highest.

\subsection{Key Limitations and Discussion Points}
Despite the strong results, several limitations remain. First, all supervised labels come from U.S. LiDAR. Although the landcover-matched validation is encouraging, true labeled validation in other regions remains an important next step. Second, the model is largely static in time and therefore inherits any staleness in the underlying feature layers. Major land-use change, rapid construction, wildfire disturbance, or canopy loss may require feature refreshes or retraining. Third, the \SI{100}{m} grid is appropriate for screening and coordination studies but may not capture sub-grid structure relevant to final engineering design at the exact terminal location.

A broader discussion point concerns the relationship between predictive accuracy and downstream propagation benefit. Our work improves the clutter input itself; a natural next step is to quantify how these improvements translate into changes in end-to-end propagation estimates, coordination contours, or siting decisions under specific regulatory workflows. That extension would make the business case even stronger by directly tying better RCH estimation to link-budget or exclusion-zone outcomes.

In this revision, we therefore add proxy domain-aware validation metrics for the clutter-height prediction itself, including tolerance-band accuracy, large-error tails, ITU-regime agreement, and physically plausible attribution patterns. A full end-to-end propagation study remains future work, because it
would require scenario-specific antenna heights, link geometry, frequency, and regulatory assumptions; however, the added metrics provide a closer connection between the regression evaluation and the RF planning decisions that RCH is intended to support.

\section{Conclusion}
We presented a globally deployable and interpretable machine learning framework for predicting Representative Clutter Height from open geospatial data. By training on LiDAR-derived labels from USGS 3DEP and leveraging a heterogeneous feature stack spanning vegetation, land cover, terrain, thermal cues, spectral reflectance, and anthropogenic intensity, the model substantially outperforms the fixed defaults used in ITU-R P.452-18. The final LightGBM model achieves \SI{1.79}{m} MAE and $R^2=0.765$ on held-out U.S. data, while interpretability analysis shows that the learned relationships are physically sensible. Importantly, the improvement is not assessed from $R^2$ alone: the model is also evaluated through operational error tolerances, asymmetric large-error behavior, ITU-regime consistency, and physically interpretable feature attributions, providing a stronger domain-grounded basis for deployment. Additional experiments demonstrate useful transferability beyond the United States, robustness outside forested environments, and limited value from more complex segmentation-derived features at the \SI{100}{m} scale.

Taken together, these results support a practical conclusion: RCH estimation can be modernized using open remote-sensing data and interpretable machine learning without sacrificing deployment realism. This creates a path toward more accurate and more scalable clutter inputs for satellite ground-station siting and interference analysis, especially in workflows where static categorical assumptions are no longer sufficient.

\section*{Acknowledgment}
The authors thank the Project Kuiper team at Amazon, particularly Yash Chandramouli, Mohammed Alvi, and Joel Mierzejewski, for proposing the research direction and providing mentorship. Their expertise in satellite ground-station planning helped shape the scope, priorities, and evaluation criteria of this work. The authors also acknowledge the University of Maryland Application Development Club for facilitating this collaboration. This study benefited from Google Earth Engine and the U.S. Geological Survey 3D Elevation Program.

\bibliographystyle{IEEEtran}
\bibliography{references}

\end{document}